%% file: icra2020_apple_dataset.tex
\documentclass[letterpaper, 10 pt, journal, twoside]{IEEEtran}  

\IEEEoverridecommandlockouts                              

\usepackage[dvipsnames]{xcolor}
\usepackage{graphicx}
\usepackage{amsmath}
\usepackage{amssymb}
\usepackage{adjustbox}
\usepackage{multirow}
\usepackage{multicol}
\usepackage{dblfloatfix}
\usepackage{makecell}
\usepackage{rotating}
\usepackage{caption}
\usepackage{subcaption}

\graphicspath{{figures/}}

\newcommand{\specialcell}[2][c]{%
	\begin{tabular}[#1]{@{}c@{}}#2\end{tabular}}

\title{MinneApple: A Benchmark Dataset for Apple Detection and Segmentation
}

\author{Nicolai H{\"a}ni$^{1}$, Pravakar Roy$^{1}$ and Volkan Isler$^{1}$%
	\thanks{*This work was supported by the USDA NIFA MIN-98-G02. The authors acknowledge the Minnesota Supercomputing Institute (MSI) at the University of Minnesota for providing resources that contributed to the research results reported within this paper.}
	\thanks{$^{1}$The authors are with the Department of Computer Science and Engineering, University of Minnesota, Minneapolis, MN, 55455, USA
		{\tt\small haeni001@umn.edu, royxx268@umn.edu, isler@umn.edu}}%
}

\usepackage{hyperref}
\begin{document}
	
\maketitle

\input{abstract}
\input{intro}
\input{relwork}
\input{image_collection}
\input{image_statistics}
\input{analysis}
\input{conclusion}

{\small
	\bibliographystyle{ieee}
	\bibliography{icra2020bib}
}

\end{document}

%% file: abstract.tex
\begin{abstract}
	In this work, we present a new dataset to advance the state-of-the-art in fruit detection, segmentation, and counting in orchard environments. While there has been significant recent interest in solving these problems, the lack of a unified dataset has made it difficult to compare results. We hope to enable direct comparisons by providing a large variety of high-resolution images acquired in orchards, together with human annotations of the fruit on trees. The fruits are labeled using polygonal masks for each object instance to aid in precise object detection, localization, and segmentation. Additionally, we provide data for patch-based counting of clustered fruits. Our dataset contains over 41'0000 annotated object instances in 1000 images. We present a detailed overview of the dataset together with baseline performance analysis for bounding box detection, segmentation, and fruit counting as well as representative results for yield estimation. We make this dataset publicly available and host a CodaLab challenge to encourage a comparison of results on a common dataset. To download the data and learn more about the MinneApple dataset, please see the project website:~\url{http://rsn.cs.umn.edu/index.php/MinneApple}. Up to date information is available online.
\end{abstract}

%% file: intro.tex
\section{Introduction}
\IEEEPARstart{D}{etection}, counting, and localization of fruits in orchards are important tasks in agricultural automation. They allow farmers to manage and optimize resources and make informed decisions during harvest. Fruit detection and localization are also precursors to automated fruit picking, which is one of the most labor-intensive processes~\cite{Calvin_economic_2010}. 

\begin{figure}[t!]
	\centering
	\begin{subfigure}[h]{0.3\linewidth}
		\centering
		\includegraphics[width=\textwidth]{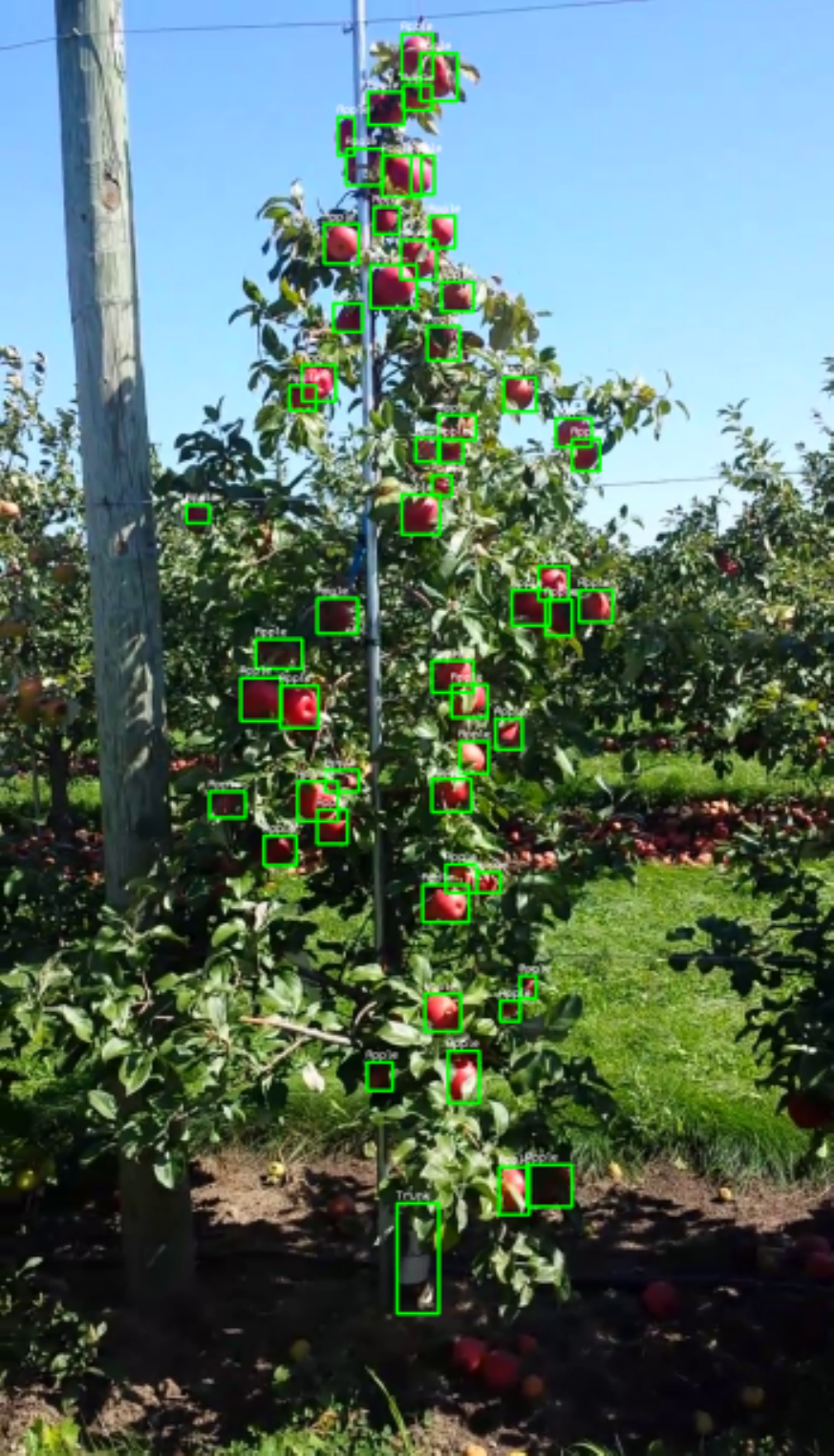}\caption{Detection}\label{fig:detection}
	\end{subfigure} \hfil
	\begin{subfigure}[h]{0.3\linewidth}
		\centering
		\includegraphics[width=\textwidth]{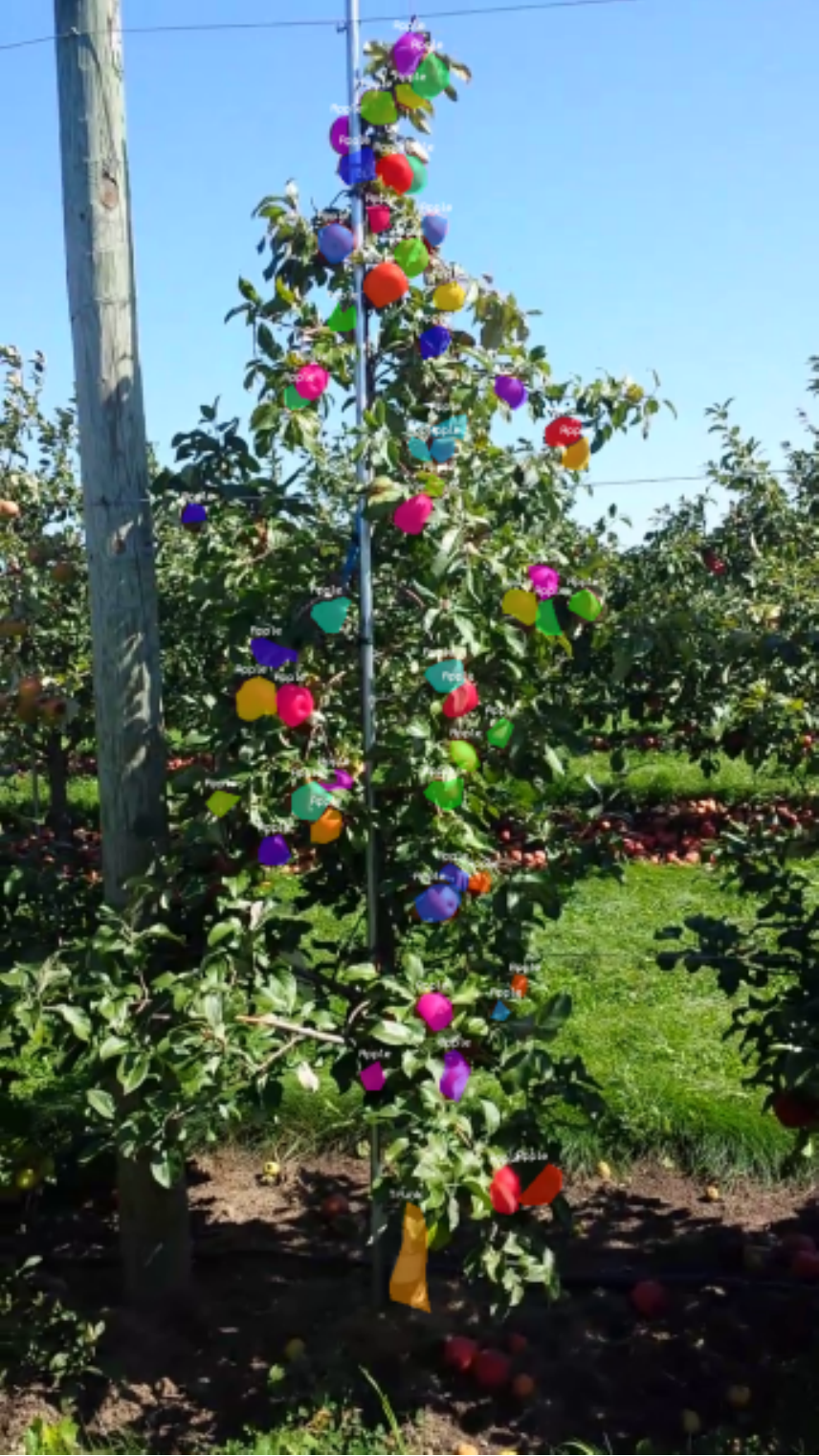}\caption{Segmentation}\label{fig:instance}
	\end{subfigure}\hfil
	\begin{subfigure}[h]{0.33\linewidth}
		\centering
		\includegraphics[width=\textwidth]{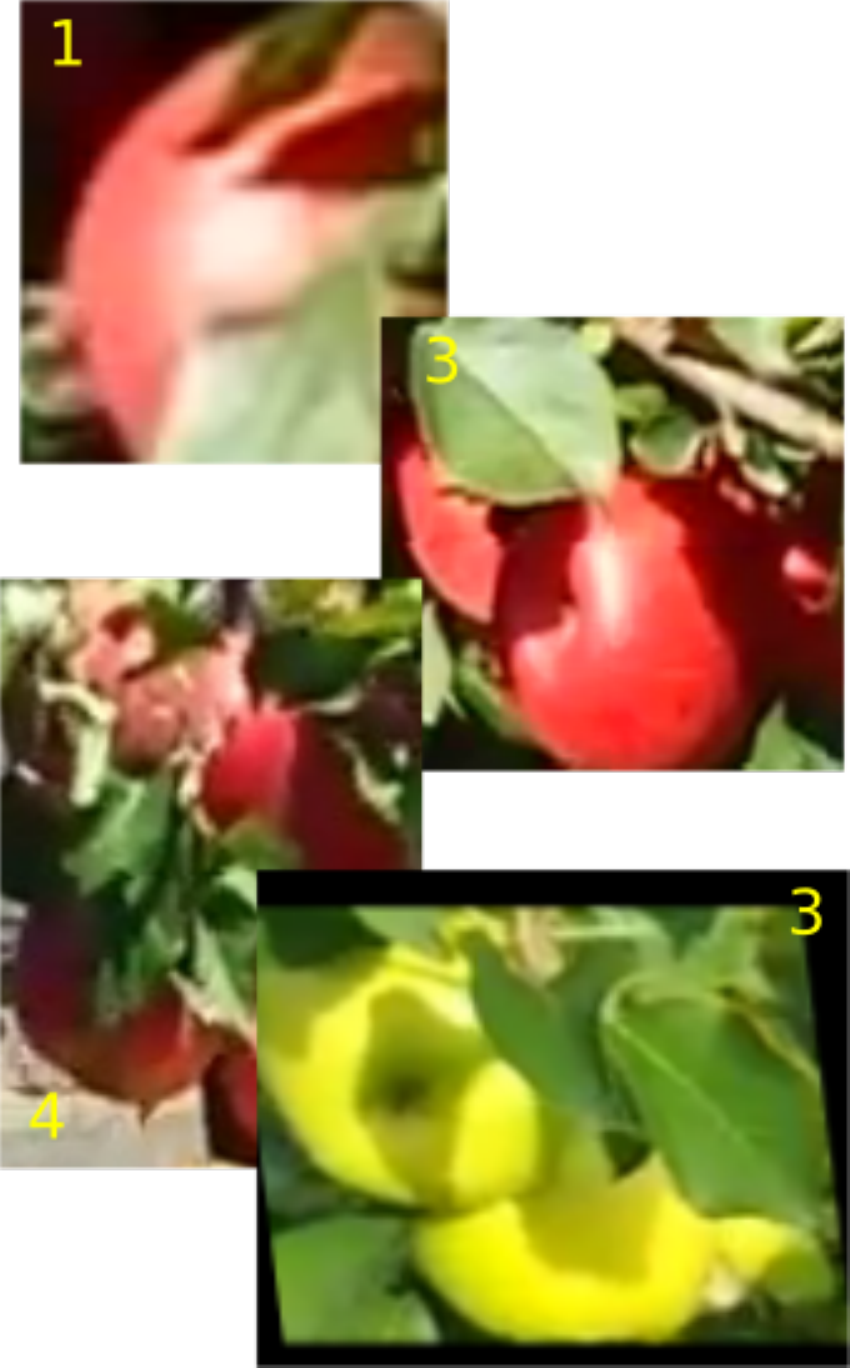}\caption{Counting}\label{fig:counting}
	\end{subfigure}
	\begin{subfigure}[h]{\linewidth}
		\centering
		\includegraphics[width=\textwidth]{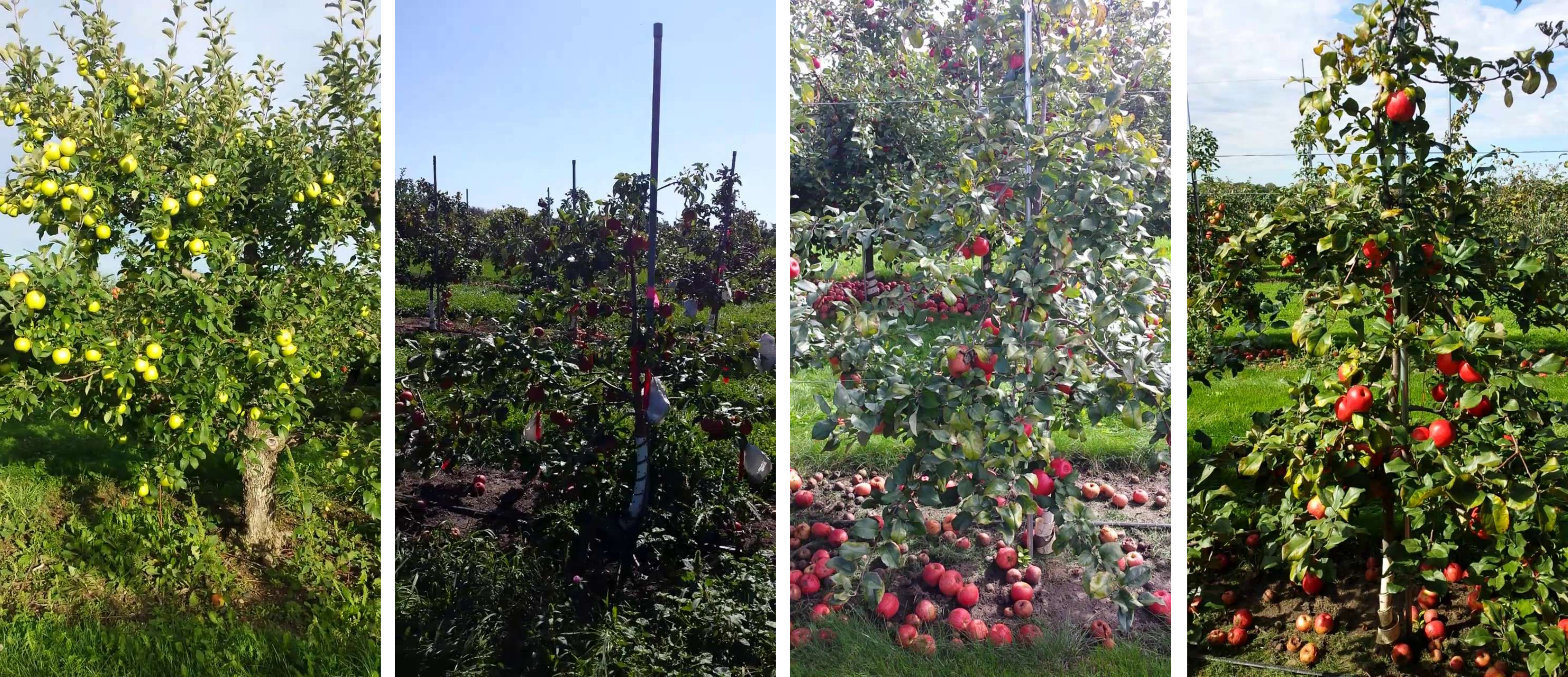}\caption{Sample images}\label{fig:variety}
	\end{subfigure}
	\caption{MinneApple contains precise semantic object instance annotations, from which one can extract bounding boxes for detection~\ref{fig:detection} and semantic labels~\ref{fig:instance}. We also provide an additional dataset to evaluate patch based counting of overlapping fruits~\ref{fig:counting}. MinneApple contains data from 17 different tree rows sporting large variety~\ref{fig:variety}.}
	\label{fig:overview}
\end{figure}

Researchers have used a variety of sensor technologies and algorithms to tackle fruit detection, but cameras together with computer vision techniques are the most common. Unfortunately, using computer vision techniques in outdoor orchard settings comes with a unique set of challenges: 1)~varying illumination conditions, 2)~varying appearances of fruits on trees, 3)~and fruit occlusion by foliage, branches or other fruits (see Figure~\ref{fig:variety}). While early detection methods primarily relied on hand-designed features~\cite{gongal_sensors_2015} recent years have seen the incorporation of deep learning methods~\cite{sa_deepfruits:_2016, bargoti_deep_2017, chen_counting_2017, roy_vision-based_2017, hani_comparative_2019}.  However, due to the lack of standardized benchmark datasets and testing metrics for precision agriculture, one can not compare these approaches directly with each other. In this paper, we introduce a new dataset and benchmark evaluation suit for apple detection, segmentation, and counting in orchard settings together with analysis of baseline algorithms. 

\begin{table*}[b!]
	\caption{Comparison of datasets used in recent research papers on apple detection and counting.} 
	\label{tab:relwork}
	\begin{center}
		\begin{tabular}{c|c c c c c c c c}
			\textbf{Fruit Detection} & type & \# train images & \# test images & \# annotations & \# scenes & resolution & ground truth & public \\
			\hline
			Bargoti and Underwood~\cite{bargoti_deep_2017} & outdoor & 729 & 112 & 5765 & 1 & $308 \times 202$ & circles & yes \\
			Stein et al.~\cite{stein_image_2016} & outdoor & 1154 & 250 & 7065 & 1 & $500 \times 500$ & circles & yes \\
			Sa et al.~\cite{sa_deepfruits:_2016} & indoor & 100 & 22 & 359 & 1 & $1296 \times 964$ & boxes & yes \\
			Liu et al.~\cite{liu_robust_2018} & outdoor & 100 & - & - & 1 & $1920 \times 1200$ & boxes & no \\
			\textbf{MinneApple (ours)} & outdoor & 670 & 331 & 41,325 & 17 & $1280 \times 720$ & polygons & yes\\
		\end{tabular}
	\end{center}
\end{table*}
\begin{table*}[b!]
	\begin{center}
		\begin{tabular}{c|c c c c c c c}
			\textbf{Fruit Counting} & type & \# train images & \# test images & \# scenes & resolution & public \\
			\hline
			Chen et al.~\cite{chen_counting_2017} & outdoor & 47 & 45 & 1 & $1280 \times 960$ & no \\ 
			Rahnemoonfar and Sheppard~\cite{maryam_rahnemoonfar_deep_2017} & synthetic & 24,000 & 2400 & 1 & $128 \times 128$& no\\
			\textbf{MinneApple (ours)} & outdoor & 64597 & 5764 & 6 & varying & yes\\
		\end{tabular}
	\end{center}
\end{table*}

Benchmark datasets have been popular and at the forefront of progress in computer vision. Many of the popular datasets in computer vision~\cite{deng_imagenet:_2009,everingham_pascal_2010,lin_microsoft_2014} contain a large number of images and categories. The COCO dataset, for example, contains a category for apples. However, an apple detector trained on this dataset will perform poorly in orchards, since the dataset was created to detect apples in general settings. These popular datasets additionally contain a small number of instances per image and contain iconic images, where the objects to be detected take up a large portion of the image, all of which hurts detection performance.

In contrast, MinneApple contains $1000$ images with over $41,000$ labeled instances of apples. The object instances are small compared to the image size, and a single image may contain between 1 and 120 objects. We collected data from multiple fruit varieties over two years, to create the largest and most diverse dataset of its kind. We hope that this dataset will provide an important stepping stone in advancing the field of precision agriculture.

The rest of the paper is organized as follows: In section~\ref{sec:relwork}, we introduce current datasets and testing methods, as well as some of the algorithms used as baselines. Then we introduce the dataset and annotation procedure in section~\ref{sec:dataset}. Section~\ref{sec:stats} contains the dataset statistics and we evaluate benchmark algorithms in section~\ref{sec:analysis}. 

%% file: relwork.tex
\section{Related Work}
\label{sec:relwork}
Many computer vision techniques rely on large datasets for training, testing, and comparing different approaches to a given problem. They not only provide the means to train and evaluate new algorithms but encourage direct comparison of results. Ultimately, they provide the means for researchers to tackle new and more challenging research problems. The ImageNet~\cite{deng_imagenet:_2009}, Pascal VOC~\cite{everingham_pascal_2010} and the COCO~\cite{lin_microsoft_2014} datasets have made millions of \emph{labeled} images available to the public and enabled breakthroughs in image classification and object segmentation. Similarly, researchers released specialized datasets for autonomous driving~\cite{cordts_cityscapes_2016, yu_bdd100k:_2018, geiger_are_2012} or pedestrian detection~\cite{ess_robust_2009, dollar_pedestrian_2009}.
While precision automation and automated yield mapping have seen much research effort~\cite{bargoti_deep_2017, stein_image_2016,roy_surveying_2016, chen_counting_2017, hani_apple_2018, hani_comparative_2019}, each of these papers used their own datasets of varying completeness and level of detail.

\subsection{Fruit detection}
The first step in a yield estimation or fruit picking pipeline is the detection of the fruit. Early methods mostly relied on static color thresholds for detection. The limitations of these methods were often compensated by adding additional sensors, such as thermal- or Near Infrared (NIR) cameras. Gongal et al.~\cite{gongal_sensors_2015} offer a comprehensive overview of these early detection methods. More recent papers used object detection networks to detect fruits~\cite{sa_deepfruits:_2016, bargoti_deep_2017,  hani_comparative_2019}. Sa et al.~\cite{sa_deepfruits:_2016} used a combination of NRI and RGB images of fruits in indoor environments. Their dataset contains only $122$ images from which training and test data are extracted. Bargoti and Underwood~\cite{bargoti_deep_2017} used a similar network for apple detection. They released their dataset of roughly $1000$ image crops that they used for training and testing. The images are of size $308 \times 202$ pixels with circular annotation of the fruits. Stein et al.~\cite{stein_image_2016} used a Faster RCNN network to detect mango fruits. \cite{chen_counting_2017} used a Fully Convolutional Network (FCN) to compute feature maps. Integrating these feature maps gives them a yield estimate. They split the dataset of $71$ images $50/50$ between training and testing. In our previous work~\cite{roy_surveying_2016, roy_vision-based_2017, roy_vision-based_2019, hani_comparative_2019} we presented results on HD sized images, showing parts of an orchard row. We presented multiple methods including semi-supervised Gaussian Mixture Model (GMM) a Faster R-CNN object detector and a semantic segmentation network. The training dataset contained $100$ images, and the test set contained $207$ images. In contrast, we have increased the size of the dataset by a factor of $\times 3.5$ for this work.

\subsection{Fruit counting}
After detection of the fruits, they need to be counted.
Rahnemoonfar and Sheppard~\cite{maryam_rahnemoonfar_deep_2017} used synthetic data to train a network to classify images according to fruit counts. They test their approach on $100$ annotated images. Chen et al.~\cite{chen_counting_2017} used a fully convolutional network together with a regression head for counting. They used a total of 71 orange- and 21 apple images from which they extracted image patches for training. Roy and Isler~\cite{roy_vision-based_2017}  proposed an unsupervised counting method based on Gaussian Mixture Models. They used a manually annotated dataset of $440$ images for testing. In our previous work~\cite{hani_apple_2018}, we used a neural network to count clustered fruits. We trained a network on $13000$ patches and tested our approach on $4$ different datasets with a total of $2800$ images. 

\begin{figure*}[b!]
	\centering
	\def\svgwidth{\textwidth}
	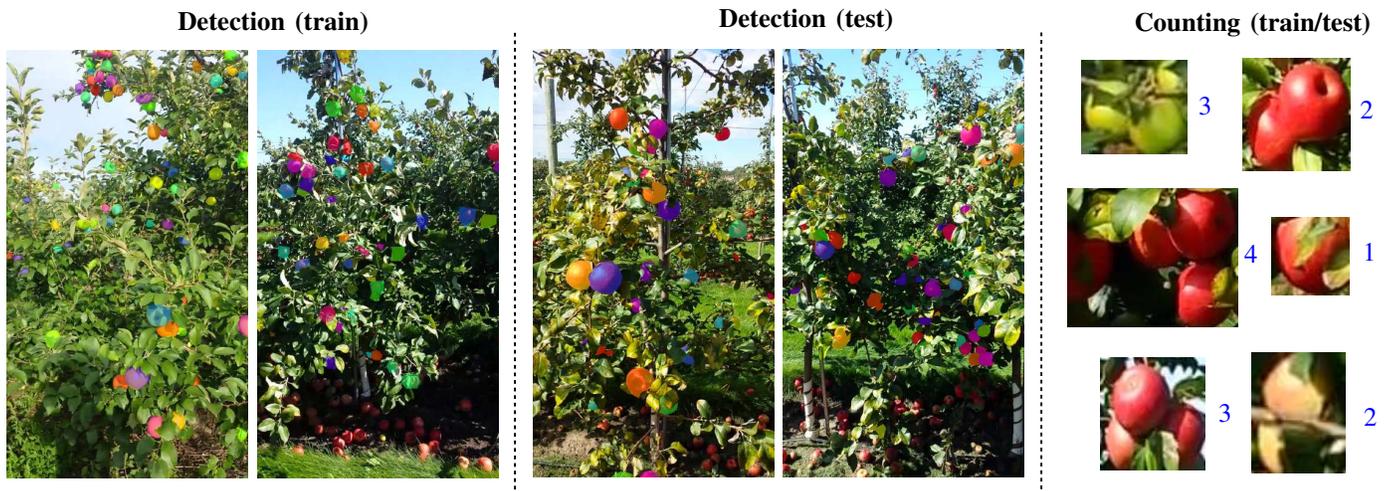
	\caption{Samples of annotated images of the detection, segmentation and counting datasets. The detection/segmentation datasets are annotated with object instance masks, while the counting dataset contains image patches and a corresponding ground truth count. }
	\label{fig:examples}
\end{figure*}

\subsection{Comparison of Datasets}
Table~\ref{tab:relwork} summarizes the problem with current fruit detection and counting datasets. Because labeling effort is time-consuming and costly, researchers have focused on small datasets with little if any in dataset variety. Acquired images are chopped into smaller chunks to increase the dataset size artificially. These crops show only a small portion of the original image, which shows in the small number of labeled fruits. While this technique increases dataset size, it does not increase dataset variation, and the developed methods are prone to overfitting. Another issue is that the whole datasets are split into training/testing. Such splits lead to in-dataset testing, which makes it impossible to analyze an algorithm on its generalization capabilities. 

The MinneApple dataset tries to correct these problems. The dataset contains only full resolution images. Data for the train/test splits are taken from different tree rows and different years.  We included a variety of apple species and illumination conditions to avoid overfitting. The MinneApple dataset gives researchers a tool to test their algorithms in an unbiased way and compare to other approaches.

%% file: figures/examples.pdf_tex
\begingroup%
  \makeatletter%
  \providecommand\color[2][]{%
    \errmessage{(Inkscape) Color is used for the text in Inkscape, but the package 'color.sty' is not loaded}%
    \renewcommand\color[2][]{}%
  }%
  \providecommand\transparent[1]{%
    \errmessage{(Inkscape) Transparency is used (non-zero) for the text in Inkscape, but the package 'transparent.sty' is not loaded}%
    \renewcommand\transparent[1]{}%
  }%
  \providecommand\rotatebox[2]{#2}%
  \newcommand*\fsize{\dimexpr\f@size pt\relax}%
  \newcommand*\lineheight[1]{\fontsize{\fsize}{#1\fsize}\selectfont}%
  \ifx\svgwidth\undefined%
    \setlength{\unitlength}{1083.40446412bp}%
    \ifx\svgscale\undefined%
      \relax%
    \else%
      \setlength{\unitlength}{\unitlength * \real{\svgscale}}%
    \fi%
  \else%
    \setlength{\unitlength}{\svgwidth}%
  \fi%
  \global\let\svgwidth\undefined%
  \global\let\svgscale\undefined%
  \makeatother%
  \begin{picture}(1,0.3512282)%
    \lineheight{1}%
    \setlength\tabcolsep{0pt}%
    \put(0.12538255,0.34187913){\color[rgb]{0,0,0}\makebox(0,0)[lt]{\lineheight{1.25}\smash{\begin{tabular}[t]{l}\textbf{Detection (train)}\end{tabular}}}}%
    \put(0.52209923,0.34366316){\color[rgb]{0,0,0}\makebox(0,0)[lt]{\lineheight{1.25}\smash{\begin{tabular}[t]{l}\textbf{Detection (test)}\end{tabular}}}}%
    \put(0.8266595,0.3413233){\color[rgb]{0,0,0}\makebox(0,0)[lt]{\lineheight{1.25}\smash{\begin{tabular}[t]{l}\textbf{Counting (train/test)}\end{tabular}}}}%
    \put(0,0){\includegraphics[width=\unitlength,page=1]{examples.pdf}}%
    \put(0.87360407,0.27953495){\color[rgb]{0,0,1}\makebox(0,0)[lt]{\lineheight{1.25}\smash{\begin{tabular}[t]{l}3\end{tabular}}}}%
    \put(0.9919678,0.27642164){\color[rgb]{0,0,1}\makebox(0,0)[lt]{\lineheight{1.25}\smash{\begin{tabular}[t]{l}2\end{tabular}}}}%
    \put(0.88809029,0.0540056){\color[rgb]{0,0,1}\makebox(0,0)[lt]{\lineheight{1.25}\smash{\begin{tabular}[t]{l}3\end{tabular}}}}%
    \put(0.99473881,0.05143038){\color[rgb]{0,0,1}\makebox(0,0)[lt]{\lineheight{1.25}\smash{\begin{tabular}[t]{l}2\end{tabular}}}}%
    \put(0.9071084,0.17081521){\color[rgb]{0,0,1}\makebox(0,0)[lt]{\lineheight{1.25}\smash{\begin{tabular}[t]{l}4\end{tabular}}}}%
    \put(0.99407304,0.17233023){\color[rgb]{0,0,1}\makebox(0,0)[lt]{\lineheight{1.25}\smash{\begin{tabular}[t]{l}1\end{tabular}}}}%
  \end{picture}%
\endgroup%

%% file: image_collection.tex
\section{Image Collection}
\label{sec:dataset}

The data for this paper were collected at the University of Minnesota's Horticultural Research Center (HRC) between June 2015 and September 2016. Since this is a university orchard, used for phenotyping research, it is home to a large variety of apple tree species. We collected video footage from different sections of the orchard using a standard Samsung Galaxy S4 cell phone. During data collection, we acquired video footage by facing the camera horizontally at a single side of a tree row and moving (by foot) along the tree row with approximately 1 m/s. Moving the camera at slow speeds mitigates motion blur effects. We then extracted every fifth image from these video sequences. For the test datasets, we extracted every 30th image.

\subsection{Detection and Segmentation Datasets}
For detection and localization of the fruits, we collected $17$ different datasets over two years, ten for training and seven for evaluation. We included fruits of different colors and at different stages of the ripening cycle. The datasets were taken either from the sunny or shady side of the tree row, and we spread out data capture over multiple days to get more varied illumination conditions. See Figure~\ref{fig:examples} for samples of the annotated images in our dataset.

\textbf{Training Sets:} We sampled ten datasets from six different tree rows for training purposes. Dataset show either the front (sunny) or back (shady) side of a tree row. From these ten datasets, we randomly selected and annotated $670$ images of resolution $1280 \times 720$ pixels. All of these datasets were acquired in $2015$ at the HRC, and they contain different apple varieties, fruits across different growing stages and a variety of tree shapes.

\textbf{Test Sets:} To evaluate detection/segmentation and yield estimation performance; we arbitrarily chose four different sections of the orchard. We collected seven videos from these four segments in 2016. Acquiring datasets during different years guarantees the independence of the test set. Additionally, we collected yield estimation ground truth for three tree rows by hand collecting and counting per tree yield and by measuring fruit diameters after harvest. Yield estimation includes additional steps, such as fruit tracking and tree row merging. Since we do not include a baseline for tracking, we provide only anecdotal results for yield estimation in this paper.

\begin{figure*}[b!]
	\centering
	\begin{subfigure}[h]{0.47\textwidth}
		\centering
		\includegraphics[width=\textwidth]{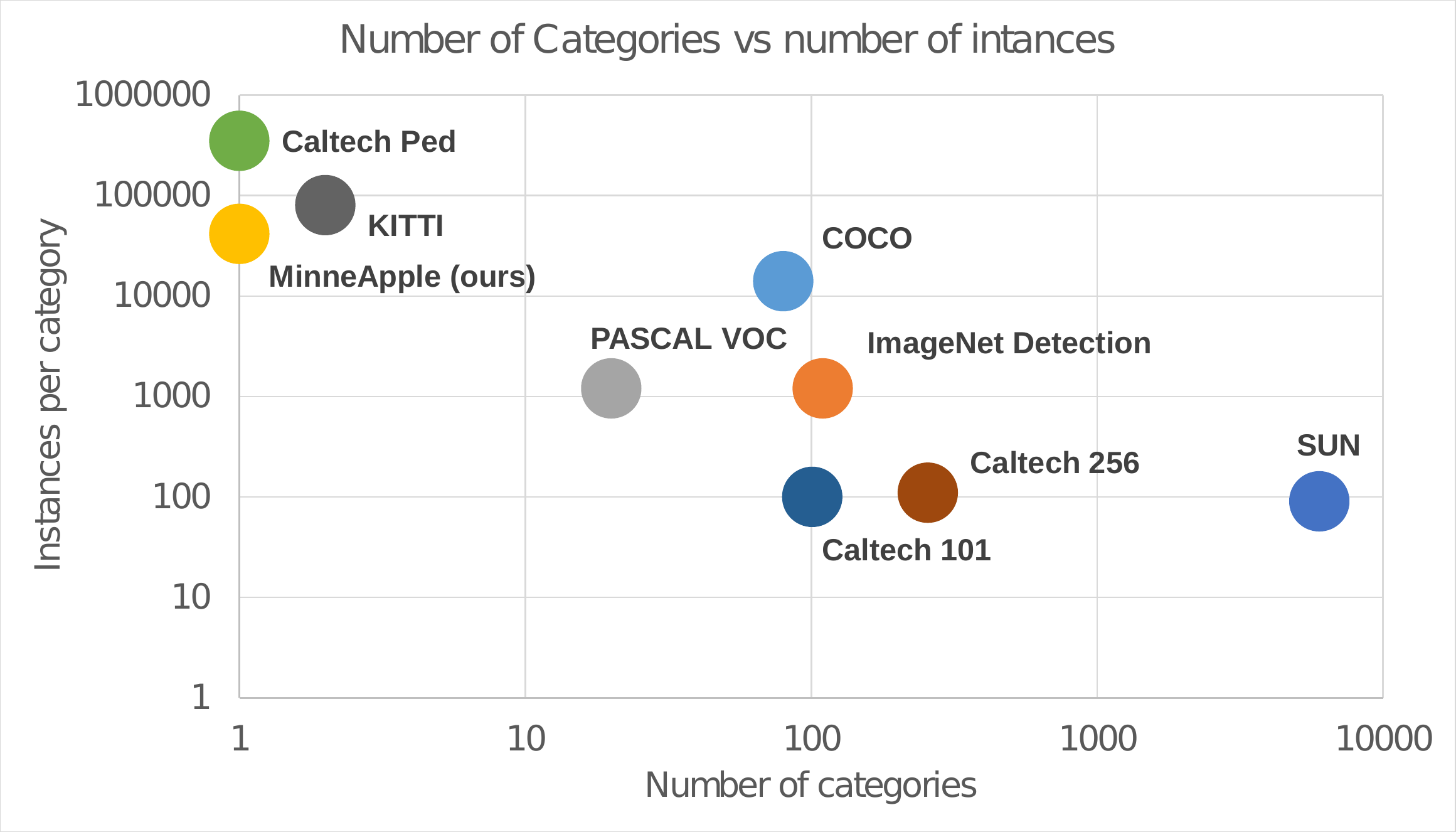}\caption{}\label{fig:categories}
	\end{subfigure} \hfil
	\begin{subfigure}[h]{0.47\textwidth}
		\centering
		\includegraphics[width=\textwidth]{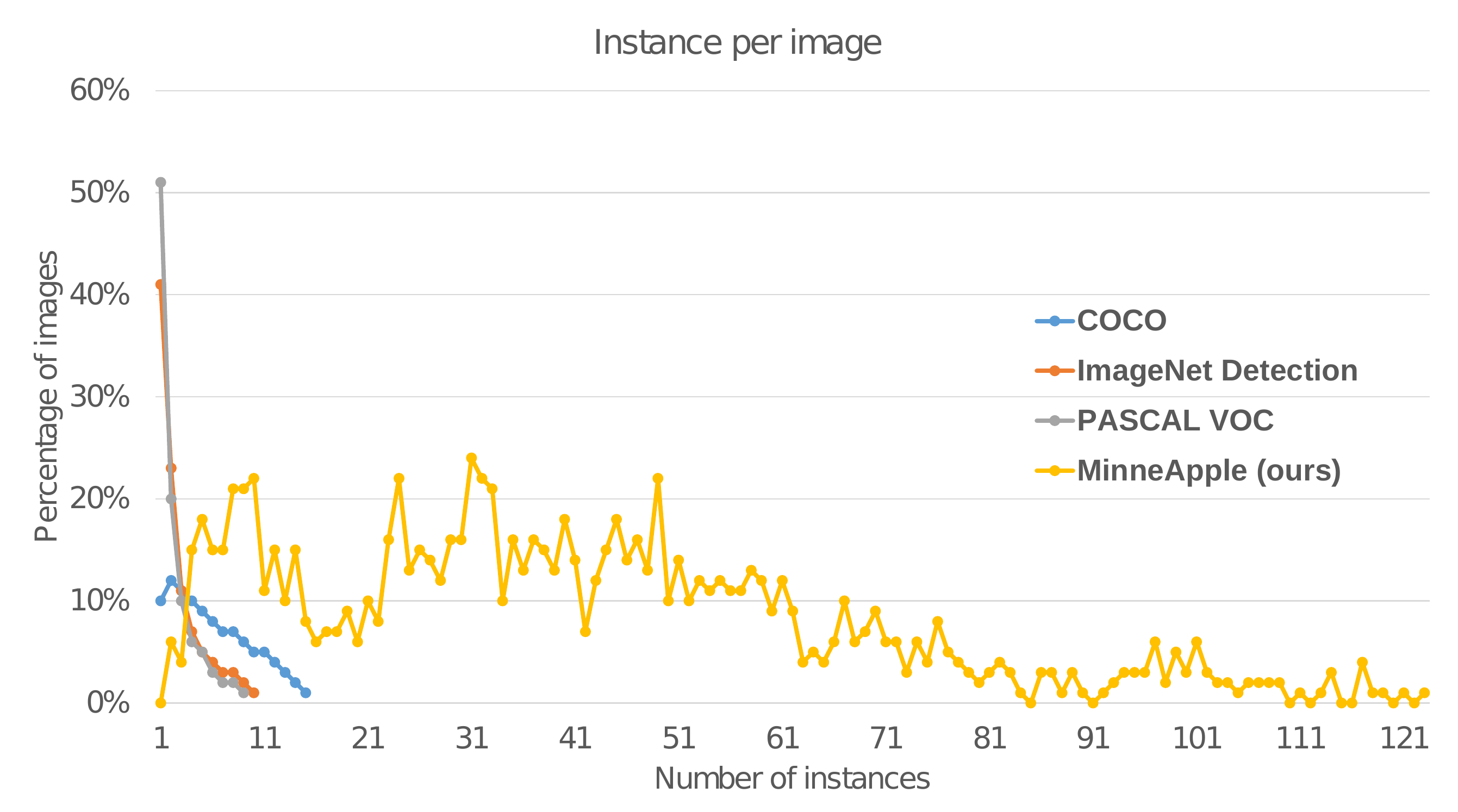}\caption{}\label{fig:instances}
	\end{subfigure}
	\caption{a) Number of annotated instances per category for some common datasets in comparison. b) Distribution of number of annotated object instances per image for COCO, ImageNet Detection, PASCAL VOC compared with MinneApple. While the other datasets contain mainly 1-5 objects, ours contains up to 120 instances per image.}
	\label{fig:stats}
\end{figure*}

\subsection{Counting Datasets}
\textbf{Training Sets: } We provide two annotated datasets to train patch-based counting approaches. One of these datasets contains green, and one contains red apples, and both were acquired in 2015. Both datasets were obtained from the sunny side of the tree row. In total, we obtained $13000$ image patches, which we annotated manually with a ground truth count. Additionally, we extracted $4500$ patches at random that do not contain apples as negative examples. See Figure~\ref{fig:examples} for samples of the annotated images in our dataset.

\textbf{Test Sets: } The test dataset consists of a total of $2874$ image patches taken from four image sequences. Two of the test datasets contain red apples, one contains greens, and one contains a mixture of colors. Additionally, we acquired the fourth dataset from a further distance to test the algorithms generalization capability for counting low-resolution fruits.

\section{Image Annotation}
\label{sec:labeling}
We next describe how we labeled images for training and evaluation. We follow the annotation method in~\cite{hani_comparative_2019}. Following established evaluation protocols, annotations for train and validation data will be released, but not for the test one. To test your algorithm please submit your results online.

\textbf{Detection and segmentation:} Fruits for the detection and segmentation datasets were annotated using the excellent VGG annotator tool~\cite{dutta_vgg_2016}. We used polygons to label fruits on trees in the foreground, while the ones on the ground and trees in the background were not tagged. Additionally, we labeled the tree trunks where visible. Please note, that we only provide annotations for fully or partially visible fruits. Each of the objects in the scene was then categorized into fruit or tree trunk. We used an internally recruited workforce for the instance labeling task. Due to the large number of instances per image, the small object size and the many occlusions of fruits instances, labeling is an arduous task.  Labeling a single image takes up to 30 minutes, which translates to roughly 18 work-hours per 1000 instances. As such, we chose to assign each image to only a single worker for labeling. Each worker is instructed in proper labeling techniques before they can begin to annotate. After the worker annotated the first ten image frames, we conducted an in-person review to give feedback and issue the first round of corrections. While the instruction and initial feedback improved annotation quality considerably, we performed an additional verification step to correct each object instance if necessary. 

\textbf{Patch-based counting: } For the patch-based counting method we used a semi-supervised GMM detector~\cite{roy_vision-based_2017} to detect image patches that are likely to contain fruits. The patches were cropped and annotated by hand with a single ground truth integer, representing the count. Due to the small resolution these patches and the large volume, the annotation task proved to be error-prone. We had two different workers annotate each image, and disparities were resolved by a third worker through a validation process.

%% file: image_statistics.tex
\section{Dataset Statistics}
\label{sec:stats}
Here, we give an overview of the properties of MinneApple in comparison to other object detection datasets. These include  COCO~\cite{lin_microsoft_2014}, ImageNet Detection~\cite{deng_imagenet:_2009} and PASCAL VOC~\cite{everingham_pascal_2010}. Each of these datasets varies considerably in the number of annotated images, image types, number of categories, number of instances per image, and the size of the annotated objects. The MS COCO dataset was created to show common objects in their natural context. The goal of the ImageNet Detection dataset was to detect a large number of object categories. PASCAL VOC contains fewer categories but focuses on objects in natural images. Our MinneApple dataset, on the other hand, focuses on detecting many small objects in highly cluttered environments.

A summary of the datasets showing the number of instances per category is shown in Figure~\ref{fig:categories}. MinneApple contains fewer categories but far more instances per category than most datasets. In this it is comparable to other specialized datasets such as the Caltech Pedestrian Detection~\cite{dollar_pedestrian_2012} dataset or the KITTI~\cite{geiger_are_2012} dataset. 
Figure~\ref{fig:instances} shows the number of instances per image in comparison to other datasets. MinneApple contains 1.5 categories and 41.2 instances on average per image. In contrast, the COCO dataset has 3.5 categories and 7.7 instances, and the ImageNet and PASCAL VOC datasets both have less than two categories and three instances per image on average. The spread of the number of instances per image compared to COCO, ImageNet, and PASCAL VOC, is also more extensive. The MinneApple dataset can contain between 1 and 120 object instances, while the other datasets have maximally 15.

\begin{figure}
	\centering
	\includegraphics[width=0.8\linewidth]{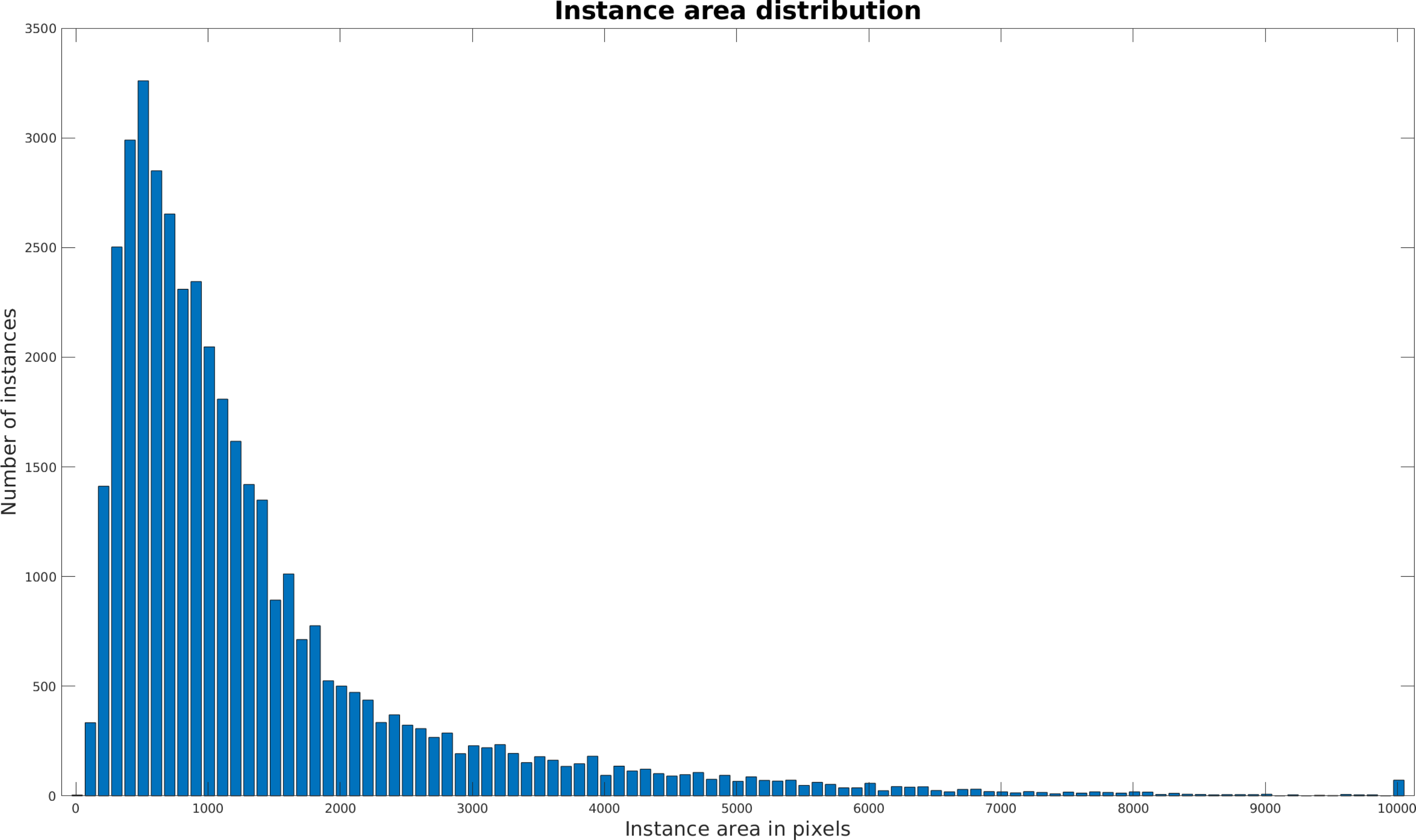}\caption{}\label{fig:area} 
	\caption{Area distribution of the objects in our dataset. The dataset contains mainly small object instances with area $<50^2$ pixels}
	\label{fig:size}
\end{figure}

\begin{table*}[b!]
	\begin{center}
		\caption{Fruit detection benchmark results for object detection approaches. Higher numbers are better and the bold marked numbers indicate the highest performing approach.}
		\label{tab:bboxdetection}
		\begin{adjustbox}{width=\textwidth}
			\begin{tabular}{c c|c c c c c c}    
				& & \multicolumn{6}{c}{Metric} \\
				& & AP [@ 0.5:0.05:0.95] & AP [@ 0.5] & AP [@ 0.75] & AP [small] & AP [middle] & AP [large] \\
				\hline
				\multirow{3}[0]{*}{\begin{sideways}Method\end{sideways}} & Tiled FRCNN~\cite{bargoti_deep_2017} & 0.341 & 0.639 & 0.339 & 0.197 & 0.519 & 0.208 \\
				& Faster RCNN~\cite{ren_faster_2015} & \textbf{0.438} & \textbf{0.775} & \textbf{0.455} & 
				\textbf{0.297} & \textbf{0.578} & \textbf{0.871} \\
				& Mask RCNN~\cite{he_mask_2017} & 0.433 & 0.763 & 0.449 & 0.295 & 0.571 & 0.809 \\
			\end{tabular}
		\end{adjustbox}
	\end{center}
\end{table*}

Finally, we analyze the average size of objects in the dataset. In general, smaller objects are harder to detect and require specialized network structures~\cite{hutchison_diagnosing_2012}.  For COCO, PASCAL VOC and ImageNet Detection, roughly 50\% of all objects occupy no more than 10\% of the image itself. The other 50\% contains objects that occupy between 10 and 100\% of the image (evenly distributed). Our MinneApple dataset contains almost exclusively small instances. The average object size is only $40 \times 40$ pixels in an image of $1280 \times 720$ pixels, making up only 0.17\% of the original image size. 

%% file: analysis.tex
\section{Algorithmic Analysis}
\label{sec:analysis}
We run a set of state-of-the-art algorithms on each the tasks of object detection, segmentation, and counting to establish a common baseline for future work. 

\subsection{Detection and Segmentation Baselines}
For the following experiments, we take a subset of 600 images from our dataset for training. The leftover 30 images are used for validation during training. We test each algorithm individually on each of the 331 test images and report average performance over the test dataset. 

\textbf{Detection evaluation metrics:} For bounding box detection, we follow established evaluation protocols used by other object detection datasets~\cite{everingham_pascal_2010, lin_microsoft_2014}. We report Average Precision (AP) as our main evaluation metric. Namely, we use AP starting at Intersection over Union (IoU) threshold 0.5 and increase it in intervals of 0.05 up to 0.95 (shorthand notation is AP@0.5:0.05:0.95).  Additionally, we provide AP@0.5 and AP@0.75. Since our dataset contains many small objects, we report AP scores for small (object area $< 32^2$ pixels), middle ($32^2 \geq \text{ object area } \geq 96^2$) and large objects (area $> 96^2$). We evaluate three different models. 

\textbf{Faster RCNN:} The latest implementation of Faster RCNN~\cite{ren_faster_2015} with a ResNet-50 backbone. The network uses pretrained COCO weights for initialization. Faster RCNN consists of a region proposal head and two branches for bounding box regression and classification. We used parameters in the paper for optimization.

\textbf{Tiled Faster RCNN:} A reimplementation of Bargoti and Underwoods~\cite{bargoti_deep_2017} proposed model. Due to memory constraints, they split the training images into $500 \times 500$ pixel chunks, with an overlap of 50 pixels. The detections of the individual chunks are aggregated and filtered using non-maximum suppression.  We added a ResNet-50 backbone, a Feature Pyramid (FPN)~\cite{lin_feature_2017} head for region proposal to the network and trained it with focal loss~\cite{lin_focal_2017}. The network was initialized with weights pretrained on COCO~\cite{lin_microsoft_2014}. We follow~\cite{bargoti_deep_2017, hani_comparative_2019} in our choice of parameters for optimization.

\textbf{Mask RCNN:} Implementation of a Mask RCNN~\cite{he_mask_2017} with a ResNet-50 backbone, pretrained on the COCO dataset. In addition to using bounding box inputs as Faster RCNN, Mask RCNN has an additional branch predicting the object instance mask. 

If we compare the bounding box detection results in Table~\ref{tab:bboxdetection}, we find that processing the image in a tiled fashion performs worse than the detectors operating on the whole image. We hypothesize that this is due to the additional filtering step at the end, where non-maximum suppression is used to filter out overlapping bounding boxes. If we compare the two state of the art object detectors, we find that Faster RCNN slightly outperforms Mask RCNN. This is somewhat unexpected since Mask RCNN has access to additional information (the instance masks). Further, we find that all the detectors struggle on smaller object instances. Future research should focus on improving the object detection performance on small and medium-sized objects to achieve significant gains in overall performance. Our findings confirm Hoiem et al.~\cite{hutchison_diagnosing_2012}, which found that object size is one of the main error factors in object detection.

\textbf{Semantic segmentation evaluation metrics:} While bounding box prediction is the method predominantly used for object detection; we recognize that there exist other methods which only achieve bounding box prediction after an additional post-processing step. These methods include mainly detection through semantic segmentation. To avoid explicit bias towards bounding box prediction methods, we introduce separate benchmark algorithms for semantic segmentation. For evaluation, we follow the established metrics used by the COCO dataset~\cite{lin_microsoft_2014}. We report Intersection over Union (IoU) as the primary challenge metric. Additionally, we report class IoU for apples, pixel accuracy, and class accuracy for apple pixels. We evaluate four different models. 

\textbf{Semi-supervised GMM:} A semi-supervised clustering method based on Gaussian Mixture Models (GMM), developed by Roy and Isler~\cite{roy_surveying_2016}. The model is pretrained on an unlabeled dataset, different from the ones contained in the train and test sets. 

\textbf{User-supervised GMM:} The same model as in the semi-supervised case. The method uses human supervision to create a single model per tree row in the test set. 

\textbf{UNet (not pretrained):} A semantic segmentation network, based on a fully convolutional network architecture~\cite{ronneberger_u-net:_2015, hani_comparative_2019}. The images in the train and test sets are split into $224 \times 224$ sized chunks, and the weights of the network are initialized randomly. 

\textbf{UNet (pretrained):} the same model as the one before, but the weights are initialized from a pretrained ImageNet network.

\begin{table}[!htpb]
	\begin{center}
		\caption{Fruit detection benchmark results for semantic segmentation approaches. Higher numbers are better and the bold marked numbers indicate the highest performing approach.}
		\label{tab:segm}
		\begin{adjustbox}{width=\linewidth}
			\begin{tabular}{c c|c c c c}    
				& & \multicolumn{4}{c}{Metric} \\
				& & IoU & Class IoU & Pixel Acc. & Class Acc.  \\
				\hline
				\multirow{8}[0]{*}{\begin{sideways}Method\end{sideways}} & \specialcell[]{Semi-supervised \\ GMM~\cite{roy_vision-based_2019}} & 0.635 & 0.341 & 0.968 & 0.455 \\
				& \specialcell[]{User-supervised \\ GMM~\cite{roy_vision-based_2019}} & 0.649 & \textbf{0.455} & 0.959 & 0.634 \\
				& \specialcell[]{UNet~\cite{hani_comparative_2019} \\ (no pretraining)} & 0.678 & 0.397 & 0.960 & 0.818 \\
				& \specialcell[]{UNet~\cite{hani_comparative_2019} \\ (pre-trained)} & \textbf{0.685} & 0.410 & \textbf{0.962} & \textbf{0.848} \\
			\end{tabular}
		\end{adjustbox}
	\end{center}
\end{table}

If we compare the average performance of all methods in Table~\ref{tab:segm}, we see that UNet with pretrained weights outperforms all others. Only in the class IoU case, the user-supervised GMM method outperforms the UNet. These results are in direct contrast to our previous work~\cite{hani_comparative_2019}, where the user-supervised GMM outperformed UNet. Keep in mind though, that for this work the amount of training data increased by almost one order of magnitude. These results indicate that the deep learning network previously underperformed due to a lack of data. We further find that using pretrained weights improves performance slightly. We hypothesize that this is due to the large variety of images found in the ImageNet dataset, which allows the network to learn more descriptive features during pretraining.

\subsection{Patch-based Fruit Counting Baselines}
Next to the benchmark dataset for object detection and segmentation, we provide a dataset for patch-based fruit counting. We report baseline results for approaches which were previously published in~\cite{hani_apple_2018} for completion. We evaluate two approaches.
\textbf{GMM:} An unsupervised method based on Gaussian Mixture Models. This method fits a mixture of Gaussians probability distribution to a previously segmented image. \textbf{CNN} This method uses a network to classify the fruits into k distinct classes. The network is based on a ResNet50~\cite{he_deep_2016} backbone, and we choose to classify six classes. Table~\ref{tab:3} shows the counting accuracy. The ResNet50 network outperforms the GMM model on all of the test sets. However, the network exhibits considerable variation in counting performance. Dataset 1 and 3 contain red and mixed apples. Dataset 4 contains red apples, but the images were acquired from further away. The best performance is achieved on test dataset 3, which contains green apples. We believe that this is the case because the green fruits show considerably less color variation than the red and mixed fruits. The GMM method performs best on test dataset 1, as this dataset is closest to the GMM training data. For an in-depth analysis and qualitative comparison, we refer the reader to~\cite{hani_apple_2018}.

\begin{table}[htbp!]
	\begin{center}
		\caption{Fruit cluster counting benchmark results.}
		\label{tab:3}
		\begin{adjustbox}{width=\columnwidth}
			\begin{tabular}{c|c|c|c|c}
				Method & Dataset 1 & Dataset 2 & Dataset 3 & Dataset 4\\
				\hline
				GMM~\cite{roy_vision-based_2017} & 88.0 \% & 81.8 \% & 77.2 \% & 76.1 \% \\
				CNN~\cite{hani_apple_2018} & \textbf{88.8} \% & \textbf{92.68 \%} & \textbf{95.1 \%} & \textbf{88.5 \%}\\
			\end{tabular}
		\end{adjustbox}
	\end{center}
\end{table}

\subsection{Yield estimation}
Fruit detection and counting are integral to solving the problem of yield estimation. However, yield estimation contains additional steps to map detections and counts to tree row yield. For one, we need to track fruits across the image sequence to avoid double counting. We address tracking fruits across images in~\cite{roy_surveying_2016, roy_vision-based_2017, roy_vision-based_2019}. Due to the planar structure of modern apple orchards, fruits can be seen from both sides of the tree row. We propose a solution to this problem in~\cite{roy_registering_2018}. Table~\ref{tab:yield} shows yield estimation results using these tracking components together with the unsupervised GMM detection method and the CNN counting. These results have been previously published in~\cite{hani_comparative_2019}, and we mention them here for completion. Using this combination of components, we achieve between 95.5 and 97.8\% accuracy with respect to the harvested ground truth.

\begin{table}[htpb!]
	\caption{Yield estimation results in terms of fruit counts.} 
	\label{tab:yield}
	\begin{center}
		\begin{adjustbox}{width=\columnwidth}
			\begin{tabular}{c|c|c|c|c|c}
				\multirow{2}{*}{} & \multirow{2}{*}{\specialcell{Harvested \\ fruit counts}} & \multicolumn{2}{c}{\specialcell[]{Merged fruit counts \\ from both sides}} & \multicolumn{2}{|c}{\specialcell[]{Summed fruit counts \\ single sides}}\\
				
				& & GMM~\cite{roy_vision-based_2019} & CNN~\cite{hani_comparative_2019} & GMM~\cite{roy_vision-based_2019} & CNN~\cite{hani_comparative_2019}\\
				\hline
				Dataset-1 & $270$ & \specialcell{$256$ \\ ($94.81\%$)} & \specialcell{\textbf{258} \\ (95.56\%)} & \specialcell{$348$ \\ ($128.89\%$)} & \specialcell{$347$ \\ ($128.52\%$)}\\				
				Dataset-2 & $274$ & \specialcell{$252$ \\ ($91.98\%$)} & \specialcell{\textbf{268} \\ \textbf{(97.81\%)}} & \specialcell{$411$ \\ ($150\%$)} & \specialcell{$405$ \\ ($147.81\%$)} \\
			
				Dataset 3 & $414$ & \specialcell{$392$ \\ ($94.68\%$)} & \specialcell{$405$ \\ $(97.83\%)$} & 
				\specialcell{\textbf{422} \\ \textbf{(101.93\%)}} & \specialcell{$430$ \\ ($103.86\%$)}\\
			\end{tabular}
		\end{adjustbox}
	\end{center}
\end{table}

%% file: conclusion.tex
\section{Conclusion}
We introduced a new dataset for detecting and segmenting apples in orchards and a second dataset for counting clustered fruits. With this collection of annotated object instances, we hope to help the advancement of object detection, segmentation, and counting of small objects in cluttered environments. In creating this dataset, we wanted to emphasize the need for a diverse and unbiased dataset, containing a large number of object instances and apple varieties between the individual tree rows. Dataset statistics and results from the baseline algorithms indicate that the images contain challenging scenarios for current state-of-the-art object detection algorithms. 

There are several promising directions for future work to improve the performance of detection and counting algorithms using this dataset. Our analysis of state-of-the-art object detectors indicates that networks can gain in accuracy by putting a broader focus on small object instances (area $<32^2$ pixels). Similarly, semantic segmentation networks may employ weighting schemes to address the class imbalance between foreground (object instances) and background pixels. We hope that our dataset will help computer vision researchers working on fruit detection. 

To download and learn more about MinneApple please see the project website:~\url{http://rsn.cs.umn.edu/index.php/MinneApple}. 

\section{Acknowledgements}
This work was supported by the USDA NIFA MIN-98-G02, and UMN MnDrive. The authors acknowledge the Minnesota Supercomputing Institute (MSI) at the University of Minnesota for providing resources that contributed to the
research results reported within this paper \url{http://www.msi.umn.edu}.